\documentclass[runningheads]{llncs}
\usepackage{graphicx}
\usepackage{url}
\usepackage{hyperref}
\usepackage{orcidlink}
\begin{document}
\title{Integrating Quantized LLMs into Robotics Systems as Edge AI to Leverage their Natural Language Processing Capabilities}
%
%
\author{Miguel Á. González-Santamarta\inst{1}\orcidlink{0000-0002-7658-8600} \and Francisco J. Rodríguez-Lera\inst{1}\orcidlink{0000-0002-8400-7079} \and David Sobrín-Hidalgo\inst{3}\orcidlink{0009-0005-7673-5921} \and Ángel Manuel Guerrero-Higueras\inst{1}\orcidlink{0000-0001-8277-0700} \and Vicente Matellán-Olivera\inst{1}\orcidlink{0000-0001-7844-9658}}
\authorrunning{Miguel Á. González-Santamarta et al.}
%
\institute{Robotics Group, University of León, León, 24006, Spain\\
\email{\{mgons,fjrodl,dsobh,agueh,vmato\}@unileon.es}}
\maketitle              
\begin{abstract}
Large Language Models (LLMs) have experienced great advancements in the last year resulting in an increase of these models in several fields to face natural language tasks. The integration of these models in robotics can also help to improve several aspects such as human-robot interaction, navigation, planning and decision-making. Therefore, this paper introduces llama\_ros, a tool designed to integrate quantized Large Language Models (LLMs) into robotic systems using ROS 2. Leveraging llama.cpp, a highly optimized runtime engine, llama\_ros enables the efficient execution of quantized LLMs as edge artificial intelligence (AI) in robotics systems with resource-constrained environments, addressing the challenges of computational efficiency and memory limitations. By deploying quantized LLMs, llama\_ros empowers robots to leverage the natural language understanding and generation for enhanced decision-making and interaction which can be paired with prompt engineering, knowledge graphs,  ontologies or other tools to improve the capabilities of autonomous robots. Additionally, this paper provides insights into some use cases of using llama\_ros for planning and explainability in robotics.
\keywords{Robotics, LLMs, Prompt Engineering, Cognitive Architectures, Planning, Knowledge Graph, HRI, Explainability, XAR}
\end{abstract}

\section{Introduction}

As the field of robotics advances, the integration of artificial intelligence (AI) technologies has become increasingly prevalent, enabling robots to perform complex tasks with greater autonomy and adaptability. Large Language Models (LLMs), such as GPT-4 \cite{openai2023gpt4}, have demonstrated remarkable capabilities in text understanding and generation, such as dialog, code and ontologies; opening up new possibilities for human-robot interaction, task planning, and decision-making. However, deploying these models in resource-constrained robotics platforms poses significant challenges, particularly in terms of computational efficiency and memory requirements.

To address these challenges, we introduce llama\_ros \cite{llama_ros_2023}, a cutting-edge tool designed to integrate quantized LLMs into robotic systems using ROS 2, the Robot Operating System \cite{macenski2022robot}. The llama\_ros tool provides an efficient solution for running quantized LLMs, leveraging the power of modern AI techniques while ensuring compatibility with the real-time constraints and resource limitations inherent in robotics applications.

At the core of llama\_ros lies llama.cpp \cite{githubGitHubGgerganovllamacpp}, a highly optimized runtime engine for executing quantized LLMs in resource-constrained environments based on CPU and GPU. By employing quantization techniques~\cite{cai2017deep,lin2016fixed,ma2024era}, llama\_ros reduces the memory usage and the computational overhead associated with running large-scale deep learning models, making them suitable for deployment on edge devices commonly found in robotic platforms.

This paper presents an overview of llama\_ros and its components, highlighting its capabilities for enabling advanced AI-powered functionalities in robotic systems. We discuss its design and explore its integration with ROS 2, demonstrating how it enables communication between LLM-based modules and other components within the robotics ecosystem. Furthermore, we showcase the practical applications of llama\_ros in various robotic scenarios, ranging from natural language understanding and generation to task planning and decision-making.


The rest of the paper is organized as follows. Section~\ref{sec:sota} reviews the background and the related works. Section~\ref{sec:llama_ros} depicts the llama\_ros tool, illustrating its components and several usage cases, while Section~\ref{sec:use_cases} presents some use cases of llam\_ros. Finally, Section~\ref{sec:conclusions} presents the conclusions and the future works.

\section{Background and Related Works}
\label{sec:sota}

Recent advancements in artificial intelligence have led to the development of Large Language Models (LLMs), sophisticated systems trained on vast text datasets to understand and generate human language. While notable progress has been made with releases like ChatGPT~\cite{openai2023gpt4} (including GPT-3.5 and GPT-4), the emergence of LLaMA~\cite{llama1} and its successor LLaMA2~\cite{llama2} in various sizes (7B, 13B, 33B, and 65B) marks a significant milestone in the LLM landscape, enabling researchers to train custom models on diverse datasets.

There are several cases of LLMs used in robotics, for instance, \cite{vemprala2023chatgpt} employs ChatGPT for a wide array of robotics tasks. Moreover, ROSGPT\_VISION \cite{benjdira2023rosgpt_vision} proposes the use of LLMs along with ontologies to command robots. On the other hand, planning is also faced with LLMs in robotics, like the case of ProgPrompt~\cite{singh2023progprompt} that enables planning through code generation with LLMs.  LLM-based planning can also be employed in multi-agent systems, for instance, the work \cite{kannan2023smart} uses LLM to create a multi-robot task plan from high-level instructions.

Therefore, the increase of using LLMs in robotics has led to the integration of them into ROS 2 \cite{macenski2022robot}. For instance, the work ROS-LLM \cite{auromix/ros-llm_2024} integrates GPT-4 into ROS 2. Another example of integrating GPT in ROS 2 is RosGpt \cite{koubaa2023rosgpt} which also leverages ontology development to convert unstructured natural language commands into structured robotic instructions. However, these tools do not allow to run LLMs locally and depend on the Internet connection that the robot must have.

The deployment of LLMs locally poses challenges due to their substantial computational demands, particularly in resource-constrained environments such as edge systems within robots. To address this issue, quantization techniques~\cite{cai2017deep,lin2016fixed} have been employed, involving the reduction of model precision from floating-point to fixed-point numbers. This process significantly decreases memory and computational requirements, rendering LLMs suitable for deployment in robots with limited memory. 

The availability of LLaMA models and quantization methods has facilitated the proliferation of LLM deployment in personal computers and edge systems, supported by tools such as llama.cpp~\cite{githubGitHubGgerganovllamacpp}. The landscape of LLMs continues to evolve rapidly with the introduction of innovative models that can be publicly accessed in Hugging Face\footnote{\url{https://huggingface.co/spaces/HuggingFaceH4/open_llm_leaderboard}}. Therefore, llama\_ros is a tool to run these new models in robotics systems locally.

\section{llama\_ros}
\label{sec:llama_ros}
The llama\_ros \cite{llama_ros_2023} tool consists of a set of public ROS 2 packages\footnote{\url{https://github.com/mgonzs13/llama_ros}} that integrated llama.cpp \cite{githubGitHubGgerganovllamacpp} into ROS 2. This integration facilitates the utilization of quantized LLMs, executed through llama.cpp, inside ROS 2 systems using a set of messages created to expose the llama.cpp functionalities.

\begin{figure*}[!ht]
\centering
\includegraphics[width=1.0\textwidth]{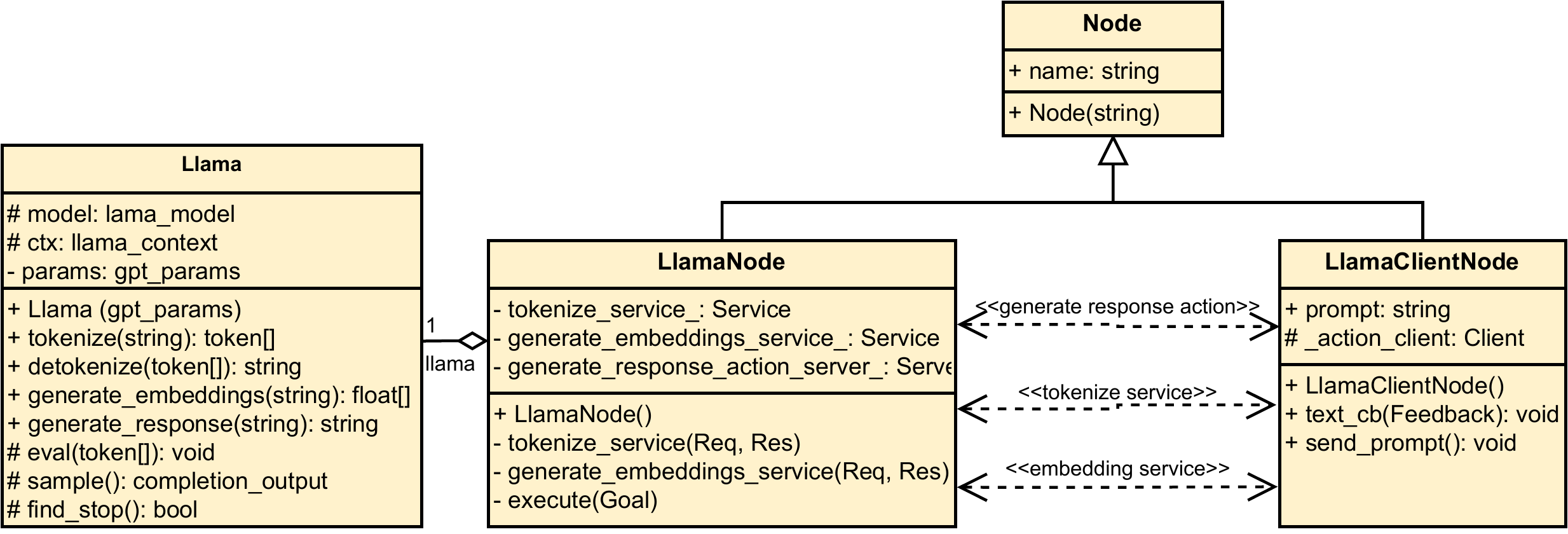}
\caption{UML diagram of llama\_ros. The diagram includes the Llama class, which wraps llama.cpp, the LlamaNode class, which presents the ROS 2 interfaces; and the LlamaClientNode, which is a node example to use llama\_ros. The ROS 2 interfaces are an action to generate a response from a prompt, a service to tokenize a text and a service to create embeddings from a text.\label{fig:llm_ros_uml} } 
\end{figure*}

The architecture of llama\_ros is depicted through the UML class diagram showcased in Figure~\ref{fig:llm_ros_uml}. At the core of this design lies the Llama class, which encompasses all the functionalities of llama.cpp independently of ROS 2 dependencies. Acting as the cornerstone, the Llama class facilitates the development of the LlamaNode class, a ROS 2 node that exposes ROS 2 interfaces, thereby facilitating the invocation of llama.cpp functions from other ROS 2 nodes. There is also an example of a node, the LlamaClienteNode, invoking these functionalities through the presented interfaces.

Therefore, the developed ROS 2 interfaces are as follows:
\begin{itemize}
    \item \textbf{Response Generation}: this ROS 2 action server is in charge of generating responses to prompts either generated by humans or other ROS 2 nodes. The action server sends feedback consisting of the next token sampled and its corresponding text. In addition, Backus-Naur notation (BNF) \cite{o2003grammatical} grammars based can be used to obligate LLMs response in different formats like JSON\footnote{\url{https://github.com/ggerganov/llama.cpp/blob/master/grammars/json.gbnf}}.

    \item \textbf{Tokenization}: this ROS 2 service performs tokenization, utilizing the tokenizer of the LLM. Tokens are the fundamental components used by LLMs. These units can vary depending on the chosen tokenization method, encompassing characters, words, subwords or other text. Tokens are assigned numerical values or identifiers and organized into sequences or vectors, serving as inputs and outputs for the model.

    \item \textbf{Embeddings}: this ROS 2 service generates embeddings from texts. These encoded representations of tokens encapsulate sentences, paragraphs, or entire documents and are generated during the prompting interaction. Positioned within a high-dimensional vector space, each dimension corresponds to a learned linguistic feature or attribute. Embeddings enable the model to discern and differentiate between different tokens or language components, facilitating language understanding and generation.
\end{itemize}

Moreover, llama\_ros is integrated into LangChain~\cite{Chase_LangChain_2022}, which is a framework employed to simplify the creation of LLM-based applications that use the new prompt engineering~\cite{sahoo2024systematic} techniques focused on structuring the prompts used in generative models. To do it, the ROS 2 interfaces of llama\_ros are wrapped in LangChain, allowing using this framework inside ROS 2 applications.

As a result, llama\_ros enables the development of LLM-based applications within edge ROS 2 systems. For instance, in human-robot interaction, llama\_ros can be utilized to enable robots to understand and generate natural language responses, facilitating more communication between humans and robots. Additionally, llama\_ros can be applied in task planning and execution, allowing robots to interpret high-level commands and execute tasks autonomously with greater precision and reliability. Thus, several applications and use cases of llama\_ros are presented in the following section.

\section{Use Cases}
\label{sec:use_cases}

In this section, we outline various applications of llama\_ros, where quantized LLMs are incorporated into robotics tasks to address diverse challenges through llama\_ros. 

\subsection{Planning and Reasoning}

Reasoning is the functionality of processing knowledge to accomplish different tasks, such as as planning and generating new knowledge. Therefore, LLMs have been used to improve the reasoning capabilities of intelligent systems. An example of this is presented in \cite{giglou2023llms4ol} which uses LLMs to create ontologies. Furthermore, LLMs have also been tested on different types of reasoning, such as deduction, induction and abduction \cite{tang2023large,bowen2024comprehensive}. This increase in using LLMs in reasoning tasks has led to the use of benchmarks, such as \cite{cobbe2021training,sakaguchi2021winogrande} to evaluate the reasoning and common sense of LLMs. 

There are also prompt engineer techniques~\cite{sahoo2024systematic} for complex reasoning capabilities such as chain-of-thought (CoT) \cite{wei2022chain}, which is based on using intermediate reasoning steps. Thus, the work of \cite{kojima2023large} has demonstrated that LLMs can work as zero-shot reasoners. The expansion of the CoT technique involves the integration of search algorithms. Consequently, \cite{yao2023tree} introduces the tree-of-thought approach, empowering LLMs to undertake deliberate decision-making processes by evaluating various reasoning paths and selecting the optimal course of action. Similarly, the graph-of-thought approach~\cite{besta2023graph} follows a comparable concept but represents potential paths in a graph structure rather than a tree.

\begin{figure*}[t]
\centering
\includegraphics[width=1.0\textwidth]{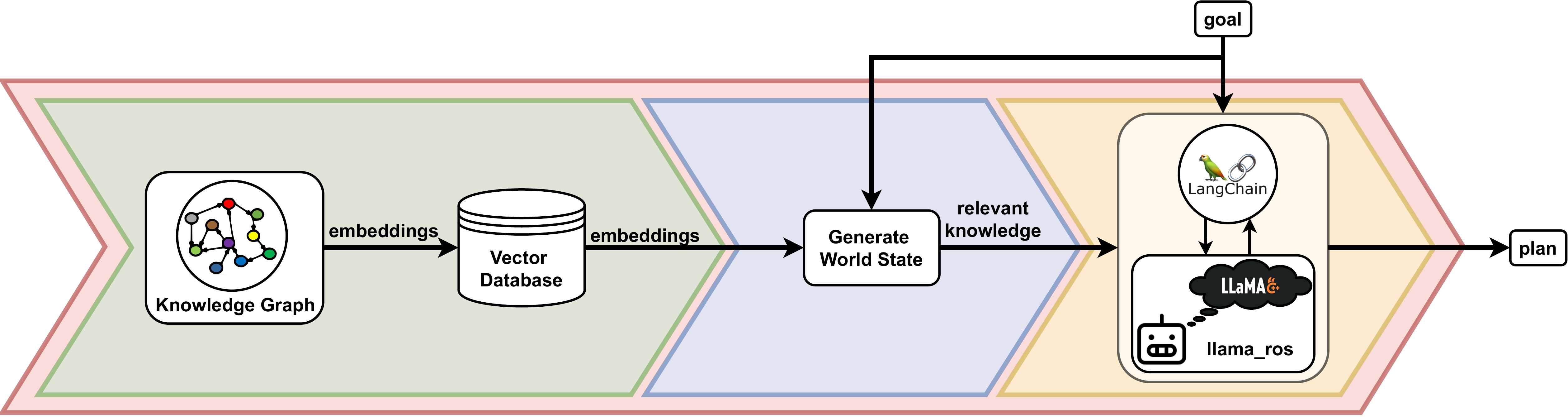}
\caption{Pipeline of the new planning functionality after integrating llama\_ros in MERLIN2. The knowledge of the robot, which is represented by a knowledge graph, is converted into embeddings and stored in a vector database. Then, applying RAG, only relevant knowledge is used to make the LLM run with llama\_ros acts as a planner.\label{fig:llm_ca} } 
\end{figure*}

In robotics, the LLM reasoning capabilities can be leveraged to carry out natural language planning. Our use case is presented  \cite{gonzálezsantamarta2023integration}. which is based on the cognitive architecture MERLIN2 \cite{GONZALEZSANTAMARTA2023100477} employed to generate behaviors in autonomous robots. MERLIN2 has a deliberative system with a symbolic knowledge base and planner. The symbolic knowledge is based on PDDL \cite{PDDL}. Thus, the knowledge base management is carried out using KANT (Knowledge mAnagemeNT) \cite{KANT}, a tool to query, create, modify and delete PDDL-based symbolic knowledge. Finally, the deliberative system can use different PDDL planners, for example, POPF \cite{popf}, to create plans to solve the high-level robot's goals.

\begin{figure}[t]
\centering
\includegraphics[width=0.7\textwidth]{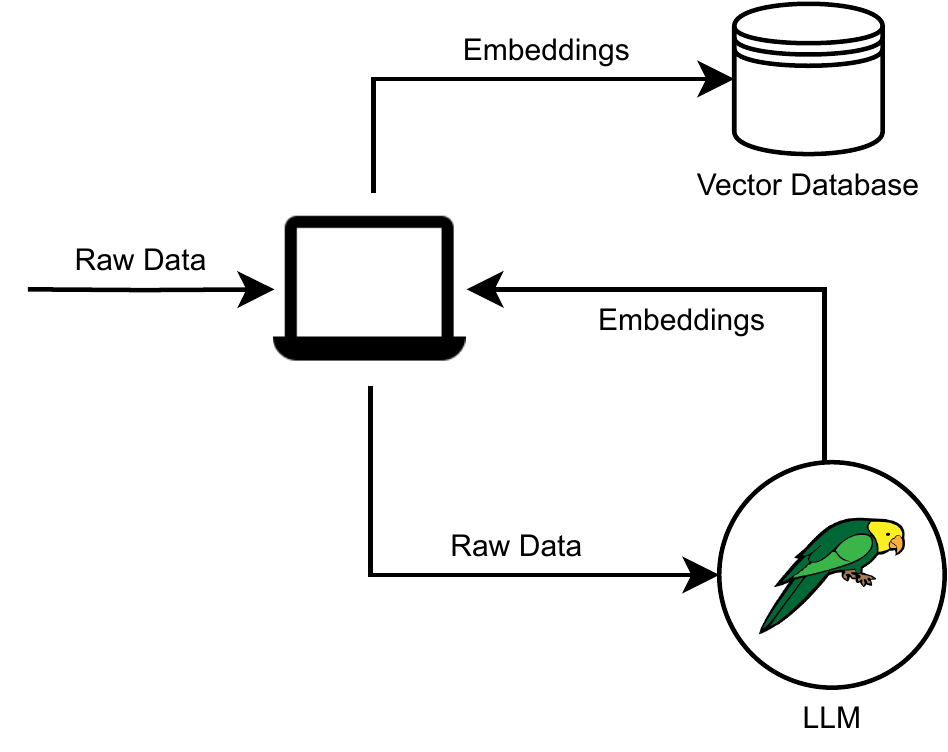}
\caption{Example of using an LLM to fill a vector database with embeddings created from texts or other data. The raw data, which could be any kind of text, such as PDF documents, are converted into embeddings, which are numerical vectors, that are stored in the vector database to be used later.\label{fig:llm_embeddings} } 
\end{figure}

Consequently, MERLIN2 is modified to replace the PDDL planners with LLMs, executed with llama\_ros, and the PDDL knowledge base with a knowledge graph. Figure \ref{fig:llm_ca} showcases the resulting planning pipeline after the modifications. The robot's knowledge, represented by a knowledge graph, is converted into embeddings stored in a vector database, as presented in Figure \ref{fig:llm_embeddings}. In this case, the knowledge of the knowledge graph, that is the knowledge that the robot has at that moment, is converted into text that corresponds with the raw data of the figure.

\begin{figure}[t]
\centering
\includegraphics[width=0.77\textwidth]{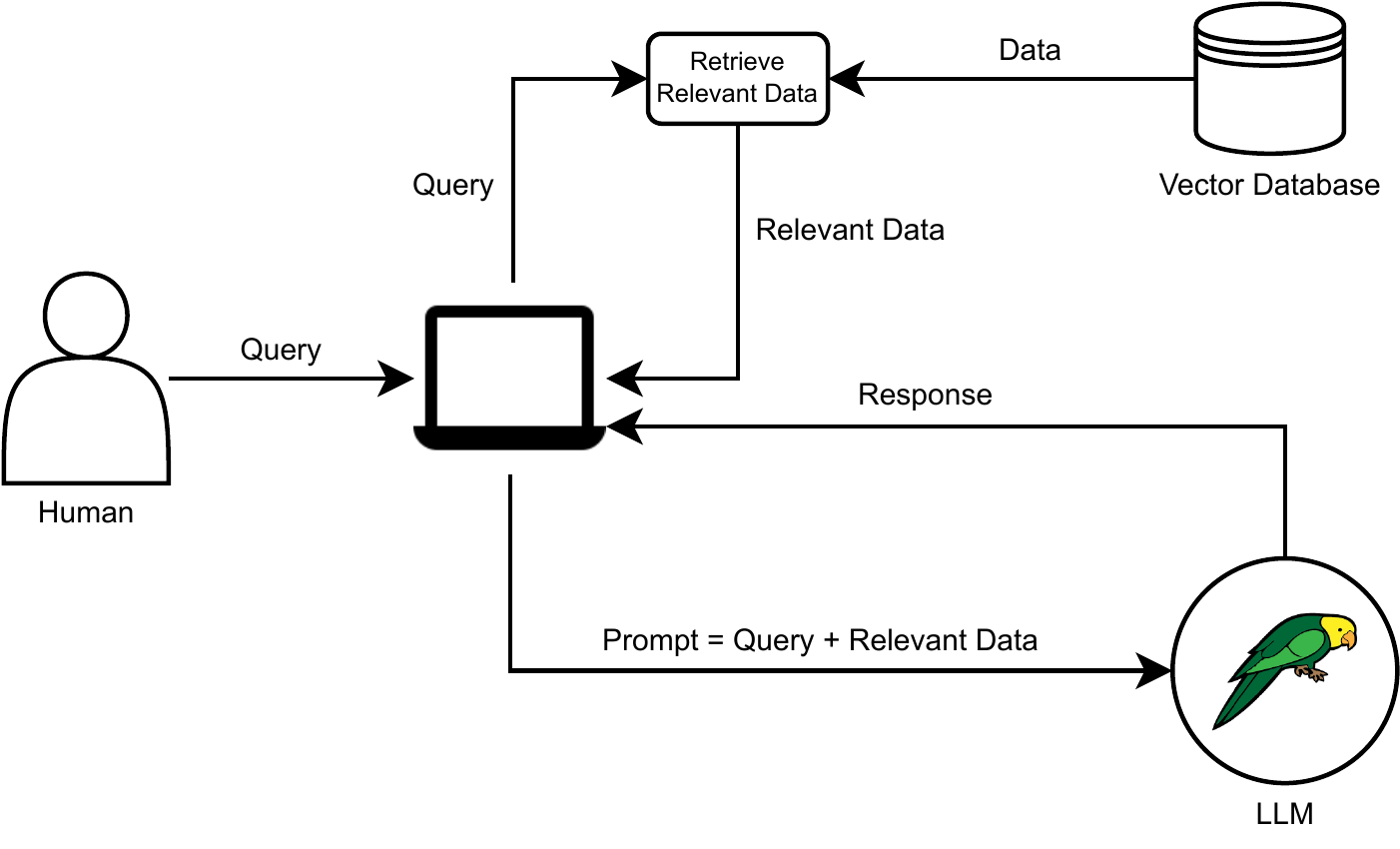}
\caption{Example of Retrieval Augmented Generation (RAG). A query is used to retrieve relevant data from the vector database. That data is used to create a refined prompt.\label{fig:llm_rag} } 
\end{figure}

Then, Retrieval Augmented Generation (RAG) \cite{lewis2020retrieval}, which is another prompt engineering technique, is applied as presented in Figure \ref{fig:llm_rag}. Using the robot's goal as a query, relevant knowledge is retrieved from the vector database. The query is also converted into embeddings which are used to search for similar vectors in the database. Then, the relevant knowledge, the goal and the actions of the robot are used to create a prompt that makes the LLM behave as a planner. Finally, the actions of the plan produced by the LLM are executed by interacting with the environment of the robot.

In this use case, llama\_ros and LangChain are used to replace the classic symbolic planner of the MERLIN2 intended to generate, control, plan and monitor behaviors in robots. RAG is also used to refine the prompt used to make the LLM plan by adding only the relevant knowledge for the specific goals. Finally, the LLM is used again after executing the plan to check if the goals are achieved in the knowledge graph of the robot.

Finally, the use of ontologies could improve the presented use case by modeling the knowledge of the knowledge graph and providing the LLM with the necessary context for LLMs, allowing them to disambiguate terms and accurately interpret the meaning of the natural language text provided to the robot by humans or from other sources such as Visual Language Models (VLM). For example, \cite{caufield2024structured} presents the SPIRES and OntoGPT for information extraction based on structured schema and standardized ontologies and LLMs. Furthermore, ontologies can provide a structured framework of knowledge, enabling the LLM to make a more informed and accurate decision.

\subsection{Explainability in Social Robots}

Human-robot interaction represents a dynamic field characterized by continuous interactions between humans and robots. Consequently, researchers in social robotics are dedicating significant efforts to enhance human trust in robots \cite{trustworthy2015}. A field that improves this trust is explainability. Thus, explainability within robotics, embodied by the concept of an eXplainable Autonomous Robot (XAR), is emerging as a prominent research domain.

In previous works \cite{gonzalez2023using}, we have experimented with the use of LLMs to explain robot behaviors from the robot's logs. Hence, our following work \cite{sobrínhidalgo2024explaining} aims to take advantage of the capabilities of LLMs in performing natural language processing tasks to generate explanations in combination with RAG to interpret data gathered from the logs of autonomous systems. The evaluation uses a navigation test from the European Robotics League (ERL), a Europe-wide social robotics competition, and a validation questionnaire to measure the quality of the explanations from the perspective of technical users.

A public repository\footnote{\url{https://github.com/Dsobh/explainable_ROS}} is available for this case. It uses llama\_ros and LangChain to fill a vector database with logs obtained from a ROS 2-based robot in runtime. Then, a human can ask natural language questions about the past events of the robot. RAG is used to take the relevant logs and create a refined prompt that the LLM uses to create the response for those questions.

\section{Conclusions}
\label{sec:conclusions}

This paper has presented the llama\_ros tool that aims to integrate and execute quantized LLMs in robotics systems as edge AI. Thanks to it, new public pre-trained or custom LLMs can be used to leverage their capabilities of natural language understanding and generation for enhanced decision-making and interaction while operating efficiently within their hardware constraints. Besides, llama\_ros is integrated into LangChain, which allows using prompt engineering techniques in robotics applications.

This way, two use cases of llama\_ros in robotics have been presented. The first one shows the use of llama\_ros and LangChain to carry out planning in a cognitive architecture. The second use case is based on using LLMs to improve the explainability degree of social robots by interpreting logs to respond to natural language questions.

Moreover, several future works can improve the llama\_ros tools. First of all, new sampling techniques can be introduced to enhance text generation. On the other hand, Visual Language Models (VLM) can also be integrated through llama.cpp. We also want to extend this work by applying llama\_ros to other uses where LLMs can be used to face robotics tasks.

Finally, the use of ontologies can improve knowledge representation as well as define the semantics of the concepts used by the LLM. This way, ontologies can model the knowledge domain while the knowledge graph stores current facts based on that domain. As a result, ontologies provide the context to allow LLMs to disambiguate terms and accurately interpret the natural language text.


\section*{Acknowledgements}
This work has been partially funded by an FPU fellowship from the Spanish Ministry of Universities (FPU21/01438); the grant PID2021-126592OB-C21 funded by MCIN/AEI/10.13039/501100011033 and by
ERDF A way of making Europe; and the Recovery, Transformation, and Resilience Plan, financed by the European Union (Next Generation) thanks to the TESCAC project (Traceability and Explainability in Autonomous Systems for improved Cybersecurity) granted by INCIBE to the University of León.


\bibliographystyle{unsrt}
\bibliography{root}

\begin{thebibliography}{10}

\bibitem{openai2023gpt4}
OpenAI.
\newblock Gpt-4 technical report.
\newblock \url{https://arxiv.org/abs/2303.08774}, 2023.

\bibitem{llama_ros_2023}
Miguel~\'{A}. Gonz\'{a}lez-Santamarta.
\newblock {llama\_ros}.
\newblock \url{https://github.com/mgonzs13/llama\_ros}, April 2023.

\bibitem{macenski2022robot}
Steven Macenski, Tully Foote, Brian Gerkey, Chris Lalancette, and William Woodall.
\newblock Robot operating system 2: Design, architecture, and uses in the wild.
\newblock {\em Science Robotics}, 7(66):eabm6074, 2022.

\bibitem{githubGitHubGgerganovllamacpp}
{G}it{H}ub - ggerganov/llama.cpp: {P}ort of {F}acebook's {L}{L}a{M}{A} model in {C}/{C}++ --- github.com.
\newblock \url{https://github.com/ggerganov/llama.cpp}, 2023.

\bibitem{cai2017deep}
Zhaowei Cai, Xiaodong He, Jian Sun, and Nuno Vasconcelos.
\newblock Deep learning with low precision by half-wave gaussian quantization.
\newblock In {\em Proceedings of the IEEE conference on computer vision and pattern recognition}, pages 5918--5926, 2017.

\bibitem{lin2016fixed}
Darryl Lin, Sachin Talathi, and Sreekanth Annapureddy.
\newblock Fixed point quantization of deep convolutional networks.
\newblock In {\em International conference on machine learning}, pages 2849--2858. PMLR, 2016.

\bibitem{ma2024era}
Shuming Ma, Hongyu Wang, Lingxiao Ma, Lei Wang, Wenhui Wang, Shaohan Huang, Li~Dong, Ruiping Wang, Jilong Xue, and Furu Wei.
\newblock The era of 1-bit llms: All large language models are in 1.58 bits.
\newblock {\em arXiv preprint arXiv:2402.17764}, 2024.

\bibitem{llama1}
Hugo Touvron, Thibaut Lavril, Gautier Izacard, and Xavier~Martinet et. al.
\newblock Llama: Open and efficient foundation language models, 2023.

\bibitem{llama2}
Hugo Touvron, Louis Martin, Kevin Stone, Peter Albert, and Amjad~Almahairi et. al.
\newblock Llama 2: Open foundation and fine-tuned chat models, 2023.

\bibitem{vemprala2023chatgpt}
Sai Vemprala, Rogerio Bonatti, Arthur Bucker, and Ashish Kapoor.
\newblock Chatgpt for robotics: Design principles and model abilities.
\newblock {\em Microsoft Auton. Syst. Robot. Res}, 2:20, 2023.

\bibitem{benjdira2023rosgpt_vision}
Bilel Benjdira, Anis Koubaa, and Anas~M Ali.
\newblock Rosgpt\_vision: Commanding robots using only language models' prompts.
\newblock {\em arXiv preprint arXiv:2308.11236}, 2023.

\bibitem{singh2023progprompt}
Ishika Singh, Valts Blukis, Arsalan Mousavian, Ankit Goyal, Danfei Xu, Jonathan Tremblay, Dieter Fox, Jesse Thomason, and Animesh Garg.
\newblock Progprompt: program generation for situated robot task planning using large language models.
\newblock {\em Autonomous Robots}, pages 1--14, 2023.

\bibitem{kannan2023smart}
Shyam~Sundar Kannan, Vishnunandan~LN Venkatesh, and Byung-Cheol Min.
\newblock Smart-llm: Smart multi-agent robot task planning using large language models.
\newblock {\em arXiv preprint arXiv:2309.10062}, 2023.

\bibitem{auromix/ros-llm_2024}
{ROS-LLM}.
\newblock \url{https://github.com/Auromix/ROS-LLM}, Apr 2024.

\bibitem{koubaa2023rosgpt}
Anis Koubaa.
\newblock Rosgpt: Next-generation human-robot interaction with chatgpt and ros.
\newblock 2023.

\bibitem{o2003grammatical}
Michael O’Neill, Conor Ryan, Michael O’Neil, and Conor Ryan.
\newblock {\em Grammatical evolution}.
\newblock Springer, 2003.

\bibitem{Chase_LangChain_2022}
Harrison Chase.
\newblock {LangChain}.
\newblock \url{https://github.com/hwchase17/langchain}, October 2022.

\bibitem{sahoo2024systematic}
Pranab Sahoo, Ayush~Kumar Singh, Sriparna Saha, Vinija Jain, Samrat Mondal, and Aman Chadha.
\newblock A systematic survey of prompt engineering in large language models: Techniques and applications.
\newblock {\em arXiv preprint arXiv:2402.07927}, 2024.

\bibitem{giglou2023llms4ol}
Hamed~Babaei Giglou, Jennifer D'Souza, and Sören Auer.
\newblock Llms4ol: Large language models for ontology learning, 2023.

\bibitem{tang2023large}
Xiaojuan Tang, Zilong Zheng, Jiaqi Li, Fanxu Meng, Song-Chun Zhu, Yitao Liang, and Muhan Zhang.
\newblock Large language models are in-context semantic reasoners rather than symbolic reasoners.
\newblock {\em arXiv preprint arXiv:2305.14825}, 2023.

\bibitem{bowen2024comprehensive}
Chen Bowen, Rune S{\ae}tre, and Yusuke Miyao.
\newblock A comprehensive evaluation of inductive reasoning capabilities and problem solving in large language models.
\newblock In {\em Findings of the Association for Computational Linguistics: EACL 2024}, pages 323--339, 2024.

\bibitem{cobbe2021training}
Karl Cobbe, Vineet Kosaraju, Mohammad Bavarian, Mark Chen, Heewoo Jun, Lukasz Kaiser, Matthias Plappert, Jerry Tworek, Jacob Hilton, Reiichiro Nakano, et~al.
\newblock Training verifiers to solve math word problems, 2021.
\newblock {\em URL https://arxiv. org/abs/2110.14168}, 2021.

\bibitem{sakaguchi2021winogrande}
Keisuke Sakaguchi, Ronan~Le Bras, Chandra Bhagavatula, and Yejin Choi.
\newblock Winogrande: An adversarial winograd schema challenge at scale.
\newblock {\em Communications of the ACM}, 64(9):99--106, 2021.

\bibitem{wei2022chain}
Jason Wei, Xuezhi Wang, Dale Schuurmans, Maarten Bosma, Fei Xia, Ed~Chi, Quoc~V Le, Denny Zhou, et~al.
\newblock Chain-of-thought prompting elicits reasoning in large language models.
\newblock {\em Advances in Neural Information Processing Systems}, 35:24824--24837, 2022.

\bibitem{kojima2023large}
Takeshi Kojima, Shixiang~Shane Gu, Machel Reid, Yutaka Matsuo, and Yusuke Iwasawa.
\newblock Large language models are zero-shot reasoners, 2023.

\bibitem{yao2023tree}
Shunyu Yao, Dian Yu, Jeffrey Zhao, Izhak Shafran, Thomas~L Griffiths, Yuan Cao, and Karthik Narasimhan.
\newblock Tree of thoughts: Deliberate problem solving with large language models.
\newblock {\em arXiv preprint arXiv:2305.10601}, 2023.

\bibitem{besta2023graph}
Maciej Besta, Nils Blach, Ales Kubicek, Robert Gerstenberger, Lukas Gianinazzi, Joanna Gajda, Tomasz Lehmann, Michal Podstawski, Hubert Niewiadomski, Piotr Nyczyk, et~al.
\newblock Graph of thoughts: Solving elaborate problems with large language models.
\newblock {\em arXiv preprint arXiv:2308.09687}, 2023.

\bibitem{gonzálezsantamarta2023integration}
Miguel~Á. González-Santamarta, Francisco~J. Rodríguez-Lera, Ángel Manuel Guerrero-Higueras, and Vicente Matellán-Olivera.
\newblock Integration of large language models within cognitive architectures for autonomous robots, 2023.

\bibitem{GONZALEZSANTAMARTA2023100477}
Miguel~Á. González-Santamarta, Francisco~J. Rodríguez-Lera, Camino Fernández-Llamas, and Vicente Matellán-Olivera.
\newblock Merlin2: Machined ros 2 planing.
\newblock {\em Software Impacts}, 15:100477, 2023.

\bibitem{PDDL}
Maria Fox and Derek Long.
\newblock {PDDL2.1: An extension to PDDL for expressing temporal planning domains}.
\newblock {\em J. Artif. Intell. Res. (JAIR)}, 20:61--124, 12 2003.

\bibitem{KANT}
Miguel~{\'A}. Gonz{\'a}lez-Santamarta, Francisco~J. Rodr{\'i}guez-Lera, Francisco Mart{\'i}n, Camino Fern{\'a}ndez, and Vicente Matell{\'a}n.
\newblock {{KANT: A Tool for Grounding and Knowledge Management}}.
\newblock In Jos{\'e}~Manuel Ferr{\'a}ndez~Vicente, Jos{\'e}~Ram{\'o}n {\'A}lvarez-S{\'a}nchez, F{\'e}lix de~la Paz~L{\'o}pez, and Hojjat Adeli, editors, {\em Bio-inspired Systems and Applications: from Robotics to Ambient Intelligence}, pages 452--461, Cham, 2022. Springer International Publishing.

\bibitem{popf}
Amanda Coles, Andrew Coles, Maria Fox, and Derek Long.
\newblock Forward-chaining partial-order planning.
\newblock pages 42--49, 01 2010.

\bibitem{lewis2020retrieval}
Patrick Lewis, Ethan Perez, Aleksandra Piktus, Fabio Petroni, Vladimir Karpukhin, Naman Goyal, Heinrich K{\"u}ttler, Mike Lewis, Wen-tau Yih, Tim Rockt{\"a}schel, et~al.
\newblock Retrieval-augmented generation for knowledge-intensive nlp tasks.
\newblock {\em Advances in Neural Information Processing Systems}, 33:9459--9474, 2020.

\bibitem{caufield2024structured}
J~Harry Caufield, Harshad Hegde, Vincent Emonet, Nomi~L Harris, Marcin~P Joachimiak, Nicolas Matentzoglu, HyeongSik Kim, Sierra Moxon, Justin~T Reese, Melissa~A Haendel, et~al.
\newblock Structured prompt interrogation and recursive extraction of semantics (spires): A method for populating knowledge bases using zero-shot learning.
\newblock {\em Bioinformatics}, 40(3):btae104, 2024.

\bibitem{trustworthy2015}
Maha Salem, Gabriella Lakatos, Farshid Amirabdollahian, and Kerstin Dautenhahn.
\newblock Towards safe and trustworthy social robots: ethical challenges and practical issues.
\newblock In {\em Social Robotics: 7th International Conference, ICSR 2015, Paris, France, October 26-30, 2015, Proceedings 7}, pages 584--593. Springer, 2015.

\bibitem{gonzalez2023using}
Miguel~A Gonz{\'a}lez-Santamarta, Laura Fern{\'a}ndez-Becerra, David Sobr{\'\i}n-Hidalgo, {\'A}ngel~Manuel Guerrero-Higueras, Irene Gonz{\'a}lez, and Francisco J~Rodr{\'\i}guez Lera.
\newblock Using large language models for interpreting autonomous robots behaviors.
\newblock In {\em Hybrid Artificial Intelligent Systems}, pages 533--544, Cham, 2023. Springer Nature Switzerland.

\bibitem{sobrínhidalgo2024explaining}
David Sobrín-Hidalgo, Miguel~A. González-Santamarta, Ángel M.~Guerrero-Higueras, Francisco~J. Rodríguez-Lera, and Vicente Matellán-Olivera.
\newblock Explaining autonomy: Enhancing human-robot interaction through explanation generation with large language models, 2024.

\end{thebibliography}

\end{document}